%% file: bare_jrnl.tex
\documentclass[journal]{IEEEtran}
\ifCLASSINFOpdf

\else

\fi

\usepackage{caption}
\usepackage{amsmath, amssymb}
\usepackage{colortbl} 
\usepackage{tabularx} 
\usepackage{xcolor}   
\usepackage{graphicx} 
\usepackage{booktabs} 
\usepackage{color}
\usepackage{mathtools}
\usepackage{hyperref}
\usepackage{cuted}

\newcommand{\floor}[1]{\lfloor #1 \rfloor}
\DeclareMathOperator*{\argmin}{arg\,min}

\begin{document}

\title{
{GBR: Generative Bundle Refinement for High-fidelity Gaussian Splatting with Enhanced Mesh Reconstruction}}

\author{Jianing~Zhang,
        Yuchao Zheng,
        Ziwei~Li,
        Qionghai~Dai,~\IEEEmembership{Fellow,~IEEE,}
        and~Xiaoyun~Yuan~\IEEEmembership{Member,~IEEE}
        
\thanks{This work is supported in part by the National Key R\&D Program of China (No. 2024YFB2809003), and in part by the National Natural Science Foundation of China (No. 62271283 and No. 62401156).
(Corresponding author: Xiaoyun Yuan.)}

\thanks{J. Zhang is with the College of future information technology, Fudan University, Shanghai 200433, China (e-mail: 22110720080@m.fudan.edu.cn). 
Y. Zheng is with the School of Biomedical Engineering, Tsinghua University, Beijing 100084, China (e-mail: zhengyc23@mails.tsinghua.edu.cn).
Z. Li is with the Key Laboratory for Information Science of Electromagnetic Waves (MoE), Fudan University, Shanghai 200433, China (e-mail: lizw@fudan.edu.cn). 
Q. Dai is with the Department of Automation, Tsinghua University, Beijing 100084, China (e-mail: qhdai@tsinghua.edu.cn).
X. Yuan is with the MoE Key Lab of Artificial Intelligence, AI Institute,
Shanghai Jiao Tong University, Shanghai 200240, China (e-mail:
yuanxiaoyun@sjtu.edu.cn).}
}

\markboth{Journal of \LaTeX\ Class Files,~Vol.~14, No.~8, August~2015}%
{Shell \MakeLowercase{\textit{et al.}}: Bare Demo of IEEEtran.cls for IEEE Journals}
\newcommand{\yuan}[1]{{\color{blue}[xiaoyun: #1]}}
\newcommand{\zhang}[2]{{\color{red}[jianing: #1]}}
\maketitle
\begin{strip}
\begin{minipage}{\textwidth}
\vspace{-110pt}
\centering
\includegraphics[width=0.95\textwidth]{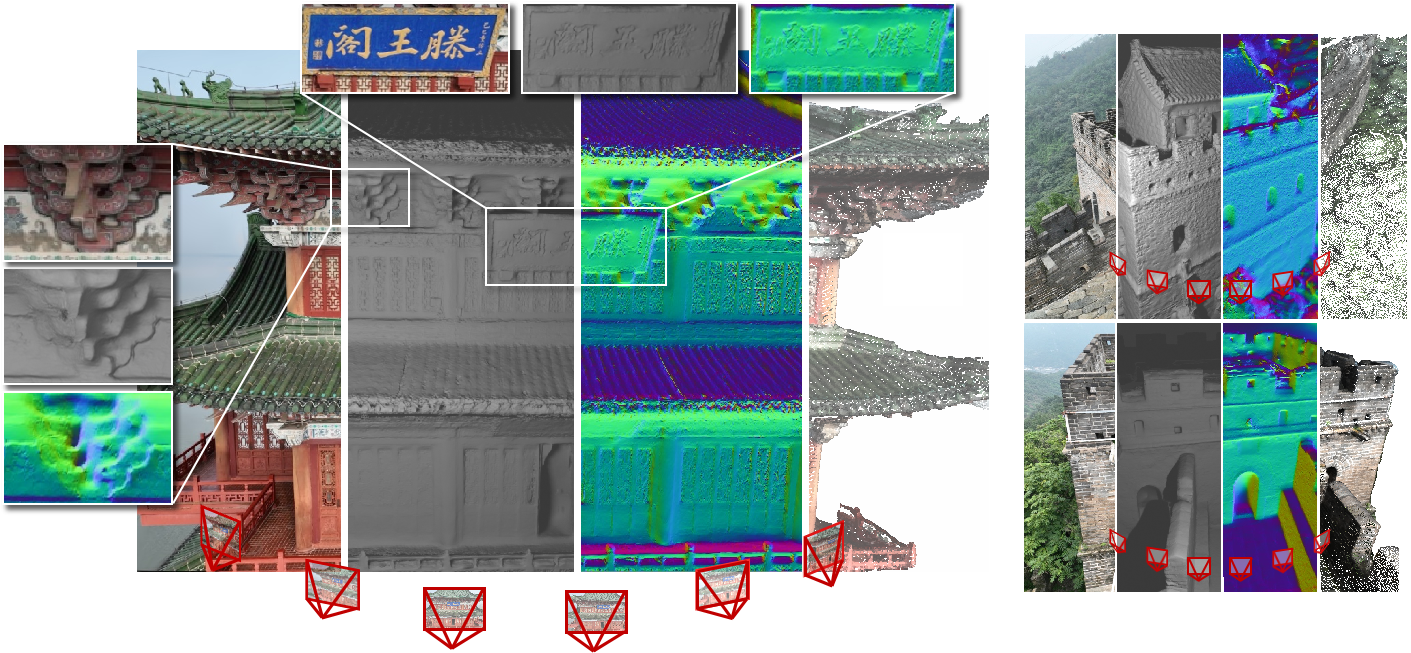}
\captionof{figure}{ Reconstruction results of GBR on large-scale real-world scenes: the Pavilion of Prince Teng and the Great Wall. The figure demonstrates our method's capability for accurate camera parameters estimation and detailed, high-fidelity surface reconstruction using \textbf{only 6 views}. From left to right, the images show novel view synthesis, mesh, normal map, and point cloud.}
\label{figurelabel}
\end{minipage}
\end{strip}

\begin{abstract}
Gaussian splatting has gained attention for its efficient representation and rendering of 3D scenes using continuous Gaussian primitives. However, it struggles with sparse-view inputs due to limited geometric and photometric information, causing ambiguities in depth, shape, and texture.
We propose GBR: Generative Bundle Refinement, a method for high-fidelity Gaussian splatting and meshing using only 4–6 input views. GBR integrates a neural bundle adjustment module to enhance geometry accuracy and a generative depth refinement module to improve geometry fidelity.
More specifically, the neural bundle adjustment module integrates a foundation network to produce initial 3D point maps and point matches from unposed images, followed by bundle adjustment optimization to improve multiview consistency and point cloud accuracy. The generative depth refinement module employs a diffusion-based strategy to enhance geometric details and fidelity while preserving the scale.
Finally, for Gaussian splatting optimization, we propose a multimodal loss function incorporating depth and normal consistency, geometric regularization, and pseudo-view supervision, providing robust guidance under sparse-view conditions. 
Experiments on widely used datasets show that GBR significantly outperforms existing methods under sparse-view inputs. Additionally, GBR demonstrates the ability to reconstruct and render large-scale real-world scenes, such as the Pavilion of Prince Teng and the Great Wall, with remarkable details using only 6 views. More results can be found on our project page \url{https://gbrnvs.github.io}.

\end{abstract}


%
\IEEEpeerreviewmaketitle

\section{Introduction}
\input{intro}
\input{relatedwork.tex}
\input{method.tex}
\input{exp.tex}


%



\ifCLASSOPTIONcaptionsoff
  \newpage
\fi



%
\bibliographystyle{IEEEtran}
\bibliography{ref}
\end{document}

%% file: intro.tex
\IEEEPARstart{G}{aussian} splatting has emerged as a promising method for achieving high-quality 3D reconstructions with lower computational and memory requirements compared to traditional voxel \cite{kutulakos2000theory} or mesh-based approaches \cite{kazhdan2006poisson}. 
By optimizing the positions, rotations, scales, and colors of the 3D Gaussian primitives and combining alpha-blending, Gaussian splatting has achieved training time in minutes and real-time rendering.
As a result, Gaussian splatting has enabled the generation of 3D content with broad applications in meta-universe\cite{deng2022fov,ye2021superplane}, autonomous driving \cite{zhou2024drivinggaussian},  robotics manipulation \cite{zheng2024gaussiangrasper}, etc.

However, existing Gaussian splatting methods still require dense-view inputs (approximately 100 views) to achieve a good performance. 
When working with sparse-view inputs (\(< 10\) views), challenges such as background collapse and the presence of excessive floaters frequently arise, leading to inaccuracies in geometry reconstruction and degraded quality in novel view synthesis (NVS).
For instance, the original Gaussian splatting pipeline\cite{Kerbl20233DGS} utilizes structure-from-motion (SfM) methods like COLMAP \cite{schonberger2016structure} to get the initial view poses and point cloud.
Unfortunately, COLMAP struggles in sparse-view inputs, as it fails to generate a sufficiently dense and complete point cloud for the initialization of Gaussian primitives, thereby limiting the geometric accuracy and mesh fidelity. 
Recent methods \cite{Fu2023COLMAPFree3G,bian2023nope}  also explored COLMAP-free initialization methods, but the dense-view inputs are still necessary. 

To realize sparse-view, high-fidelity Gaussian splatting and meshing, we have to solve the following challenges: 
1) \textbf{Geometry accuracy}. Conventional methods typically rely on Structure-from-Motion (SfM) to generate initial point clouds and view poses, achieving high accuracy but facing significant challenges with sparse-view inputs. Recent approaches \cite{fan2024instantsplat,yu2024lmgaussianboostsparseview3d} address this issue by leveraging large vision models; however, the absence of explicit multi-view constraints introduces errors in geometry and view pose estimation. These inaccuracies hinder subsequent Gaussian splatting optimization, resulting in misaligned primitives and rendering artifacts.
2) \textbf{Mesh fidelity}. Additionally, high-accuracy point clouds alone are insufficient for reconstructing high-fidelity meshes. Geometric details are often lost during Gaussian primitives optimization due to the limited multi-view information available from sparse-view inputs. While some methods \cite{xiong2023sparsegs, zhu2025fsgs} attempt to incorporate monocular depth information, inherent depth scale ambiguity compromises both geometry accuracy and fidelity.
3) \textbf{Insufficient Supervision}. Sparse-view inputs provide limited supervision for Gaussian primitives optimization, often causing the process to converge to local minima. To address this, it is essential to design effective loss functions and regularization terms that can better guide the optimization.

In this paper, we propose GBR: Generative Bundle Refinement, an effective framework to overcome the challenges mentioned above. 
%
Firstly, we propose neural bundle adjustment, which combines the widely used bundle adjustment optimizer with the neural network-based geometry estimator, such as DUSt3R\cite{wang2024dust3r}.
The DUSt3R network can directly produce dense 3D point maps from 2D image pairs,  and the conventional bundle adjustment optimizer enhances point map and view pose accuracy by incorporating explicit multi-view constraints. This combination effectively addresses the geometry accuracy challenge.
Secondly, we introduce generative depth refinement, leveraging a diffusion model to incorporate high-resolution RGB information into the 3D point map. This process ensures scale-consistent integration, maintaining depth-scale accuracy while enhancing geometric detail and smoothness.
Finally, we design a multimodal loss function incorporating confidence-aware depth loss, structure-aware normal loss, cross-view geometric consistency, and pseudo-view generation. Together, these components provide stronger supervision for Gaussian primitives optimization, leading to more accurate and robust reconstruction and rendering results. Our main contributions are as follows:

\begin{itemize}
  \item We introduce a \textbf{comprehensive Gaussian splatting pipeline} that supports camera parameters recovery, depth/normal map estimation, novel view synthesis, and mesh reconstruction while only sparse-view inputs are required. 
  \item We introduce \textbf{neural bundle adjustment}, which combines a deep neural network-based point map estimator with conventional bundle adjustment optimization to significantly enhance the accuracy of estimated camera parameters and the 3D point cloud.
  \item We propose \textbf{generative depth refinement} which uses a diffusion model to enrich geometric details while preserving depth scale, thereby boosting the fidelity of mesh reconstruction.
  \item We design a \textbf{multimodal loss function} that strengthens supervision, enabling more accurate and robust Gaussian primitives optimization.
\end{itemize}

%% file: relatedwork.tex
\section{Related Work}
\subsection{3D Representations for Novel View Synthesis}
Novel view synthesis (NVS) aims to create unseen perspectives of a scene based on a limited set of input images, a problem that has garnered substantial research interest \cite{mildenhall2019local}. NeRF \cite{mildenhall2021nerf} represents a prominent solution for photorealistic rendering by leveraging multilayer perceptrons (MLPs) to model 3D scenes, with spatial coordinates and view directions as inputs, and applying volume rendering techniques to synthesize images. While NeRF is renowned for producing high-quality renderings, it remains computationally intensive, requiring considerable training and inference time. Recent research has aimed to enhance either the quality \cite{barron2022mip,barron2023zip} or efficiency \cite{mueller2022instant,10574847} of NeRF; however, achieving an optimal balance between these aspects remains challenging.

As an alternative approach, 3D Gaussian splatting (3DGS) \cite{Kerbl20233DGS} addresses this dual objective by utilizing anisotropic 3D Gaussians \cite{zwicker2001ewa} combined with differentiable splatting techniques, resulting in both efficient and high-quality reconstructions of complex scenes. However, 3DGS necessitates careful parameter tuning, particularly in adaptive density control, which is crucial for converting sparse SfM point clouds into dense 3D Gaussian distributions. Enhancing the quality and density of SfM results may further improve the effectiveness of 3DGS, potentially accelerating its optimization process.

\subsection{3D Representations for Surface Reconstruction}
Surface reconstruction is a classic problem in 3D vision, aimed at reconstructing high-quality 3D assets from visual data. Methods like MVSNet \cite{yao2018mvsnet} utilize multi-view stereo to generate depth maps and point clouds, enabling effective surface reconstruction through traditional geometric representations. 
SurRF \cite{zhang2021surrf} introduces a surface radiance field defined on a continuous 2D surface, achieving compact representation while maintaining realistic texture and high geometric resolution. Building on the success of NVS in 3D representation, methods like NeRF and 3DGS have also been adapted for surface reconstruction tasks. NeRF-based approaches for surface modeling have incorporated occupancy fields \cite{niemeyer2020differentiable} and signed distance fields \cite{wang2021neus,yariv2021volume} to capture surface geometry. Although effective, the reliance of NeRF on MLP layers can be a bottleneck for both inference speed and flexibility in representation. To mitigate these limitations, recent approaches are moving away from MLP-heavy architectures, instead using discrete structures like points \cite{xu2022point,9915626} and voxels \cite{li2022vox,li2023neuralangelo,liu2020neural} to represent scene information more efficiently.

SuGaR \cite{guedon2024sugar} introduces a technique for extracting meshes from 3DGS by employing regularization terms to better align Gaussians with scene surfaces. This method samples 3D points from the Gaussian density field and applies Poisson surface reconstruction for meshing. While this improves geometric fidelity, the irregular shapes of 3D Gaussians can complicate smooth surface modeling, and the unordered structure of Gaussians makes the image reconstruction loss vulnerable to overfitting, potentially leading to incomplete geometry and surface misalignment issues.
To enhance multi-view consistency, methods such as 2DGS \cite{huang20242d} compress the 3D volume into planar Gaussian disks, while GOF \cite{Yu2024GOF} uses a Gaussian opacity field to extract geometry through level sets directly. PGSR \cite{chen2024pgsr} transforms Gaussian shapes into planar forms, optimizing them for realistic surface representation and simplifying parameter calculations, such as normals and distance measures. Nonetheless, achieving consistent geometry across multiple views remains a challenge in 3DGS approaches.

\subsection{Unconstrained 3D Representations}
Both NeRF and 3DGS typically require extensive data from densely captured images and often depend on SfM preprocessing, such as with COLMAP \cite{schonberger2016structure}, to estimate camera parameters and create initial sparse point clouds. This reliance on dense image captures and computationally heavy SfM limits their practical application, as errors in SfM can propagate through the 3D model, especially when input images lack sufficient overlap or exhibit low texture.

Several methods have been proposed to tackle the challenge of novel view synthesis (NVS) from sparse-input views. 
NeRF-based approaches often introduce regularization to compensate for missing visual information and improve reconstruction quality. RegNeRF \cite{niemeyer2022regnerf} enforces geometric consistency through ray sampling annealing and normalizing flow, while DGRNeRF \cite{10493053} integrates point cloud priors to refine depth accuracy, making it more robust to sparse inputs. SparseNeRF \cite{wang2023sparsenerf} further enhances depth estimation by incorporating local depth ranking and spatial continuity constraints, reducing reconstruction artifacts in under-constrained regions. FreeNeRF \cite{yang2023freenerf} enhances sparse-view synthesis by applying frequency regularization, which suppresses overfitting to incomplete observations and promotes more stable reconstructions.

Beyond regularization, some methods incorporate external guidance to infer missing details in sparse-view scenarios. DietNeRF \cite{jain2021putting} utilizes CLIP-based semantic consistency to enforce high-level feature alignment across different viewpoints, thereby improving scene coherence. ReconFusion \cite{wu2024reconfusion} employs a diffusion prior to refine both geometric and textural fidelity, helping NeRF reconstruct plausible details even from a limited number of input images. Other approaches leverage feature information to enhance NeRF's adaptability to sparse settings. PixelNeRF \cite{yu2021pixelnerf} uses convolutional feature encodings to learn scene priors, enabling generalization across multiple scenes with minimal input views. MVSNeRF \cite{mvsnerf} employs plane-swept cost volumes to provide geometry-aware feature information, allowing for efficient radiance field reconstruction from just three views while ensuring strong scene generalization ability.

Beyond NeRF, 3D Gaussian Splatting-based methods have recently received attention due to their efficiency and rendering quality in sparse-view settings. GeoRGS \cite{10643132} regularizes the growth of 3D Gaussian points to mitigate early-stage errors, which is crucial in sparse-input settings where incorrect initial geometry can lead to severe overfitting. By guiding Gaussian expansion with seed patches and enforcing depth similarity, GeoRGS improves scene consistency and reduces artifacts. SparseGS \cite{xiong2023sparsegs} incorporates depth priors and generative constraints to prevent common artifacts caused by sparse-input views such as background collapse. FSGS \cite{zhu2025fsgs} enables real-time, photo-realistic rendering by utilizing Gaussian unpooling and monocular depth estimation, requiring as few as three input views. DNGaussian \cite{li2024dngaussian} further enhances efficiency by introducing an adaptive depth regularization strategy that targetedly constrains Gaussian parameters rather than applying global depth supervision. This approach improves geometry learning while reducing artifacts, achieving a 25× faster training speed and over 3000× faster rendering compared to previous methods.

Despite these advancements, most methods still assume known camera poses, often derived from dense captures or precomputed estimates. In unconstrained settings, where pose uncertainty is prevalent, further research is needed to enhance robustness against incomplete or ambiguous camera parameters.
%
Recently, 
pose-free methods are proposed to jointly optimize camera parameters and the 3D model. Strategies like NeRFmm \cite{wang2021nerf} and BARF \cite{lin2021barf} adopt different approaches for pose encoding, enabling robust pose estimation and novel view synthesis. NeRFmm treats camera parameters as learnable variables optimized through photometric reconstruction, while BARF formulates the problem as joint scene representation learning and camera registration, leveraging coarse-to-fine positional encoding to mitigate misalignment during training.
Nope-NeRF \cite{bian2023nope}, LuNeRF \cite{cheng2023lu}, and CF-3DGS \cite{Fu2023COLMAPFree3G} utilize depth priors from monocular depth estimation to iteratively refine camera poses. By optimizing poses frame by frame with depth constraints, these methods effectively refine camera poses without relying on pre-computed initialization, making pose-free training feasible.
However, these methods typically rely on densely captured video sequences as input, making them less effective in handling sparse-view scenarios.
Methods like InstantSplat \cite{fan2024instantsplat} and LM-Gaussian \cite{yu2024lmgaussianboostsparseview3d} utilize DUSt3R \cite{wang2024dust3r} as an initializer to enable the the estimation of camera parameters and a relatively dense point cloud for training 3D Gaussian Splatting (3DGS). Similarly, ViewCrafter \cite{yu2024viewcrafter} and ReconX \cite{liu2024reconx} leverage the dense initialization generated by DUSt3R and employ video diffusion techniques to synthesize high-quality multi-view images.
However, the noise inherent in DUSt3R's initialization significantly affects geometric precision, allowing only visually appealing novel view synthesis rather than accurate geometry reconstruction.

In summary, achieving high-fidelity novel view synthesis and accurate 3D reconstruction in unconstrained scenarios, where camera poses are unknown and viewpoints are sparse, remains a significant challenge. The limited observations lead to ambiguities in both geometric estimation and 3D representation optimization, requiring effective solutions to improve reconstruction quality.

%% file: method.tex
\begin{figure*}[t]
    \centering
    \includegraphics[width=0.9\textwidth]{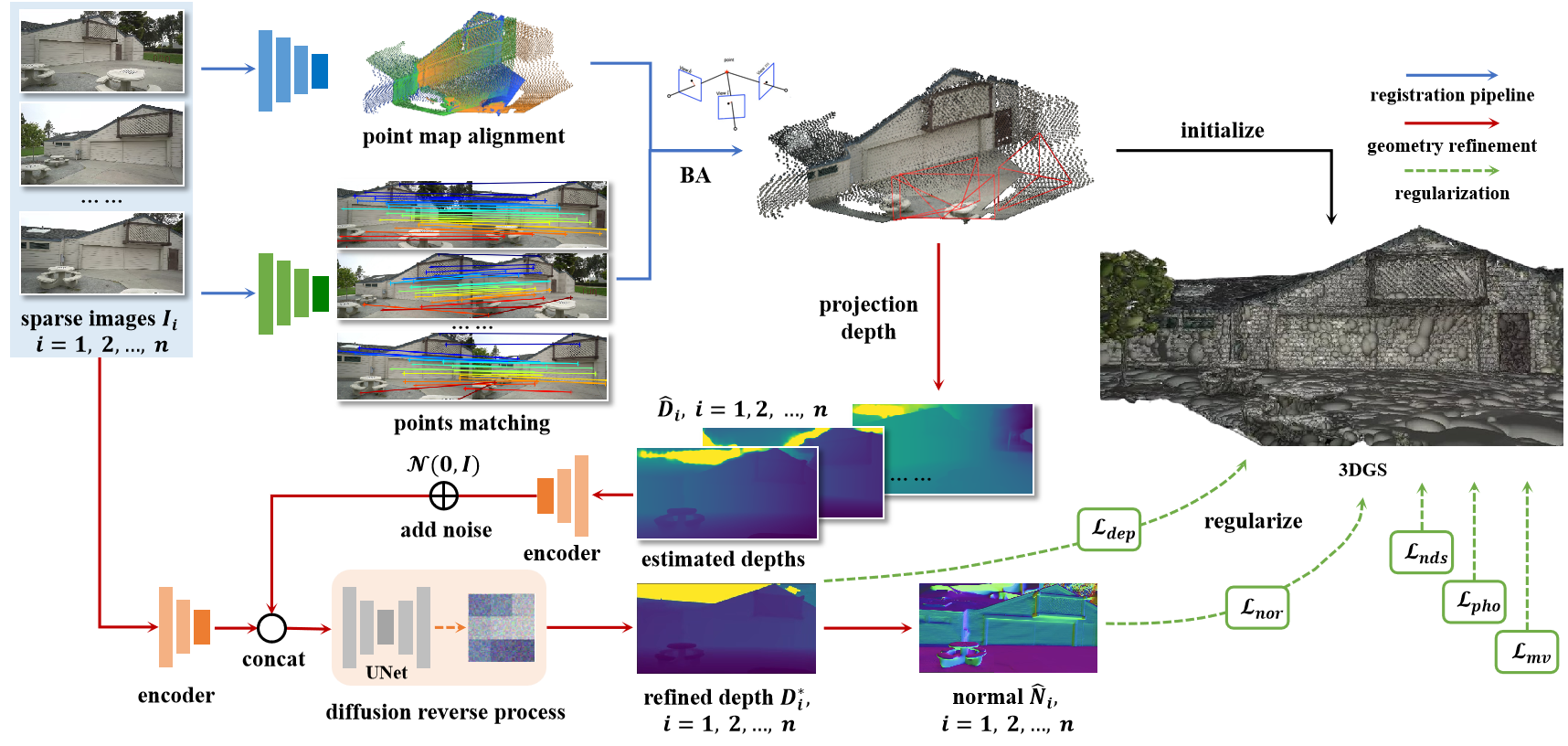} 
    \caption{Overview of our algorithm: Given unposed sparse-view inputs, we first employ neural Bundle Adjustment (neural-BA) to obtain a dense, accurate point cloud along with accurate camera parameters, providing a robust initialization for Gaussian splatting. Next, the dense point cloud is projected to obtain depth maps with accurate scale, and generative depth refinement is applied to enhance the depth details, resulting in scale-consistent, detail-rich depth and normal maps. Finally, a multimodal loss function incorporating depth, normal, pseudo-view synthesis, geometric consistency, and photometric loss is applied to optimize Gaussian splatting.
}
    \label{fig:large-image}
\end{figure*}
\section{Preliminary on 3D Gaussian Splatting}

The 3DGS algorithm reconstructs a scene by representing it as a collection of 3D Gaussian distributions, denoted \( \{G_i\} \). Each Gaussian \( G_i \) is defined by the probability density function:
\begin{equation}
G_i(x | \mu_i, \Sigma_i) = \exp \left( -\frac{1}{2} (x - \mu_i)^\top \Sigma_i^{-1} (x - \mu_i) \right),
\end{equation}
\noindent where \( \mu_i \in \mathbb{R}^3 \) represents the mean position of a point \( p_i \) in the scene, and \( \Sigma_i \in \mathbb{R}^{3 \times 3} \) is the covariance matrix defining the spatial extent of the Gaussian. The covariance \( \Sigma_i \) can be factorized into a scaling matrix \( S_i \in \mathbb{R}^{3 \times 3} \) and a rotation matrix \( R_i \in \mathbb{R}^{3 \times 3} \) as follows:
\begin{equation}
\Sigma_i = R_i S_i S_i^\top R_i^\top.
\end{equation}
\noindent This representation allows efficient rendering of the Gaussians via alpha blending. For each Gaussian, a transformation to the camera coordinate system and subsequent projection onto the 2D image plane are applied. With a transformation matrix \( W \) and the intrinsic camera matrix \( K \), the parameters \( \mu_i \) and \( \Sigma_i \) are transformed as:
\begin{equation}
\mu'_i = K W [\mu_i, 1]^\top, \quad \Sigma'_i = J W \Sigma_i W^\top J^\top,
\end{equation}
\noindent where \( J \) is the Jacobian matrix approximating the affine projection transformation. 






In order to create an image from a particular viewpoint, the color assigned to each pixel \( p \) is computed by blending the top \( K \) ordered Gaussians \( \{ G_i \mid i = 1, \cdots, K \} \) that intersect with \( p \), according to the blending equation:
\begin{equation}
c(p) = \sum_{i=1}^{K} c_i \alpha_i \prod_{j=1}^{i-1} (1 - \alpha_j),
\end{equation}
\noindent In this context, $\alpha_i$ is calculated by evaluating ${G}_i(u|\mu'_i, \Sigma'_i)$  multiplied by an opacity value learned for \( G_i \). The variable \( c_i \) represents the color associated with \( G_i \) that can be learned. The Gaussians that intersect with \( p \) are arranged based on their depth from the current viewpoint. By leveraging differentiable rendering methods, it is possible to optimize all Gaussian parameters in an end-to-end manner.

The normal map from the current viewpoint is generated via alpha blending as:
\begin{equation}
    \mathbf{N} = \sum_{i \in N} R_c^\top \mathbf{n}_i \alpha_i \prod_{j=1}^{i-1} (1 - \alpha_j),
\end{equation}
\noindent where $N$ is the number of all pixels and \( R_c \) is the rotation matrix from the camera frame to the global coordinate system and the Gaussian primitive normal vector $\mathbf{n}_i$  represents the direction of the minimum scale factor. The distance $d_i$ of a point $p_i$ from each Gaussian’s plane to the camera is computed as:
\begin{equation}
    d_i = \left( R_c^\top (\boldsymbol{\mu}_i - \mathbf{T}_c) \right) R_c^\top \mathbf{n}_i^\top,
\end{equation}
\noindent where \( \mathbf{T}_c \) denotes the camera center in world coordinates, and \( \boldsymbol{\mu}_i \) is the center of Gaussian \( G_i \). Using alpha blending, the cumulative distance map from the current viewpoint is given by:

\begin{equation}
    \mathcal{D} = \sum_{i \in N} d_i \alpha_i \prod_{j=1}^{i-1} (1 - \alpha_j).
\end{equation}

Following the depth calculation approach introduced in PGSR, once the distance and normal maps are rendered, the corresponding depth map can be derived by intersecting each pixel’s viewing ray with the Gaussian plane. The depth calculation is given by:
\begin{equation}
    \mathcal{D}(\mathbf{p}) = \frac{\mathcal{D}}{\mathbf{N}(\mathbf{p}) K^{-1} \tilde{\mathbf{p}}},
\end{equation}
\noindent where \( \mathbf{p} = [u, v]^\top \) represents the 2D coordinates of a pixel on the image plane, \( \tilde{\mathbf{p}} \) is the homogeneous representation of \( \mathbf{p} \), and \( K \) denotes the camera's intrinsic matrix.

\section{Method}
Gaussian splatting and meshing from sparse-view inputs encounter significant challenges, including geometry accuracy, mesh fidelity and limited supervision.
To tackle these challenges, we introduce GBR: Generative Bundle Refinement, a framework designed to enhance Gaussian splatting and meshing through the following key components:
1) A neural bundle adjustment module, which combines the widely used bundle adjustment optimizer with the neural network-based geometry estimator, aiming to address the geometry accuracy challenge.
2) A generative depth refinement module that employs a diffusion model to integrate high-resolution RGB information into the point cloud, enhancing geometric details and smoothness.
3) A multimodal loss function, combining depth, normal, geometric consistency, synthesized pseudo-view supervision, and photometric losses, enhances reconstruction quality.
\begin{figure}[ht]
    \centering
    \includegraphics[width=0.4\textwidth]{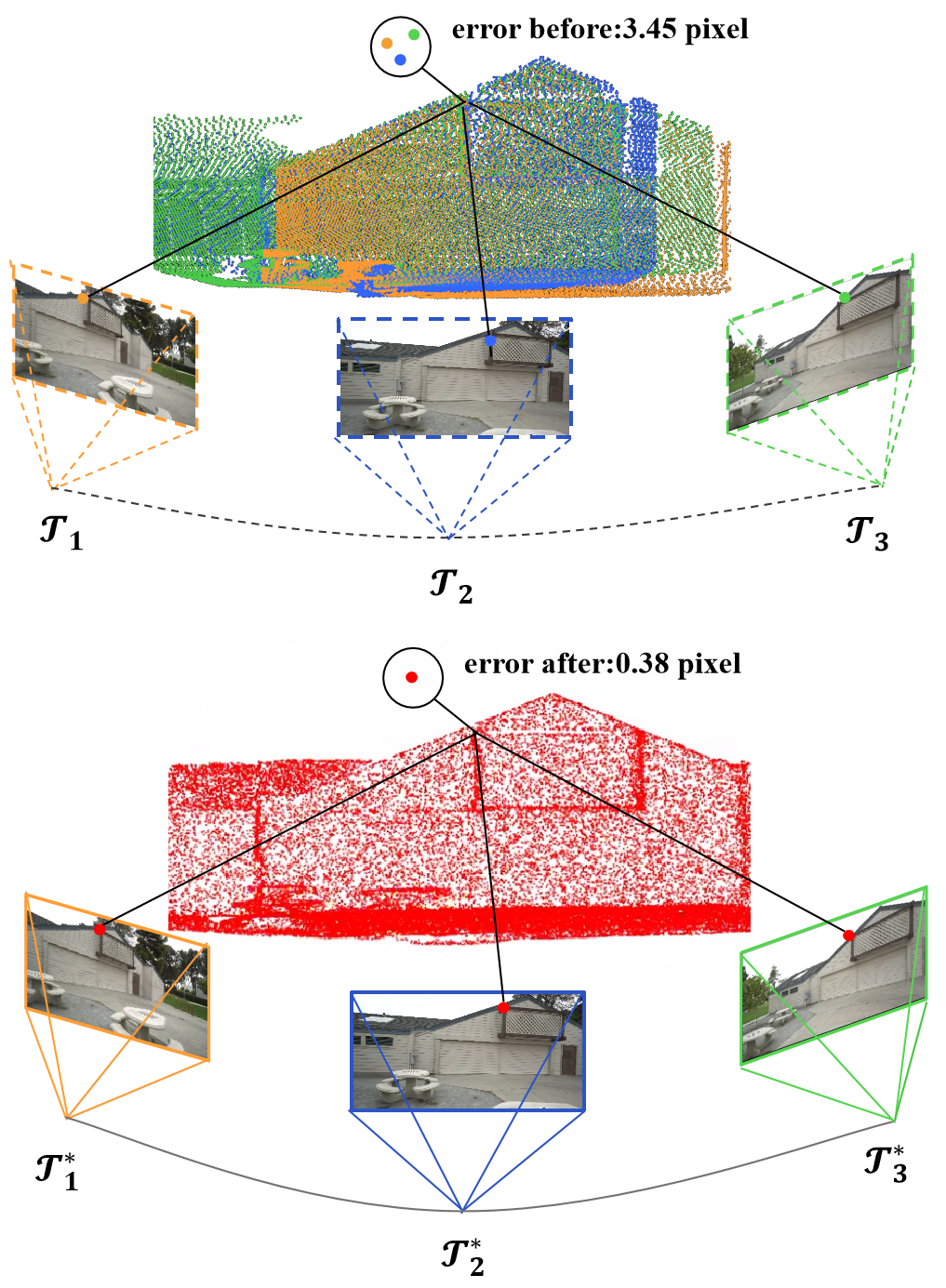} 
    \caption{Neural bundle adjustment illustration. After optimization, the point maps from different image views are well aligned.}
    \label{fig:large-image}
\end{figure}

\subsection{Neural Bundle Adjustment}
Conventional multi-view reconstruction methods depend on bundle adjustment (BA)-based SfM and dense-input images to estimate camera poses and produce an initial point cloud. 
While for sparse-view inputs, SfM struggles to estimate accurate camera poses. The produced point cloud is also usually too sparse to support Gaussian splatting initialization and training. 
In recent years, neural network-based methods have provided a preliminary solution under sparse-view inputs. However, without explicit geometry constraints, these approaches cannot guarantee the accuracy of camera poses and point clouds.
Hence, we propose the neural bundle adjustment that combines the widely used bundle adjustment optimizer with the neural network-based geometry estimator to improve the geometry accuracy and point cloud density under sparse-view inputs.
It consists of the following 3 steps: (1) Neural dense initialization; (2) Bundle adjustment refinement with dual filtering; (3) Semantic-based local refinement.

\subsubsection{Neural dense initialization}
DUSt3R (Dense and Unconstrained Stereo 3D Reconstruction) leverages transformer-based architectures to unify stereo depth estimation into a single neural network. It enables direct 3D point cloud reconstruction from unposed image pairs \(\{I_1, I_2\}\) as follows:
\begin{equation}
P^{W \times H \times 3}, C^{W \times H} = \text{DUSt3R}(I_1, I_2),
\end{equation}
where \(W\) and \(H\) represent the width and height of the input images. The output includes \(P \in \mathbb{R}^{W \times H \times 3}\), a 3D point map of \(I_1\) with each pixel encoding its corresponding 3D position, and \(C \in \mathbb{R}^{W \times H}\), a confidence map that quantifies the reliability of the reconstructed points. 

The relative poses between images can be estimated through optimizing the following registration problem:
\begin{equation}
    \argmin_{T^{k,l}, \sigma^{k,l}} \sum_{k \in K} \sum_{l \in K \setminus \{k\}} C^k \cdot C^l \|P^k - \sigma^{k,l} T^{k,l} P^l\|
\end{equation}
where \( \{P^k, P^l\} \) and \( \{C^k, C^l\} \) denote the point maps and confidence maps of the image pair \( \{I^k, I^l\} \). The purpose is to solve the transformation matrix \( T^{k,l} \) and scaling factor \( \sigma^{k,l} \) that aligns the point map from \( P^l \) to \( P^k \). This process is applied across all image pairs \( \{I^k, I^l\} \) to transform all the point maps to the same coordinate, generating a unified dense point cloud.
Additionally, the focal length \( f \) of the image \( I \) (if not pre-calibrated) can be calculated by solving the following minimization problem:
\begin{equation}
\argmin_f \sum_{i=0}^{W} \sum_{j=0}^{H} C_{ij} \left\| (i-\frac{W}{2}, j-\frac{H}{2}) - f \cdot \frac{(P_{i,j,x}, P_{i,j,y})}{P_{i,j,z}} \right\|
\end{equation}
where \( C_{ij} \) is the confidence value at pixel position \( (i, j) \); and \( P_{i,j,x} \), \( P_{i,j,y} \), \( P_{i,j,z} \) denote the x, y, and z components of the 3D point at pixel position \( (i, j) \).

Although the DUSt3R network generates an initial dense point cloud along with camera parameters, it does not explicitly account for epipolar geometric constraints. This oversight reduces the accuracy of the point cloud and the associated 3D-2D mappings, creating challenges for subsequent Gaussian splatting learning. 

\subsubsection{Bundle adjustment refinement with dual filtering}
Therefore, we propose the bundle adjustment (BA) optimization with dual filtering to introduce the epipolar geometry constraints. 
Before BA, matching points can be obtained from the DUSt3R results:
\begin{equation}
\begin{array}{rl}
\mathcal{M}_{k,l} & = \left\{ (i, j) \;\middle|\; i = \mathrm{NN}_k^{k,l}(j) \text{ and } j = \mathrm{NN}_l^{l,k}(i) \right\},
\\
\mathrm{NN}_k^{n,m}(i) & = \underset{j \in \{0, \dots, WH\}}{\operatorname{arg\,min}} \left\| P_j^{n,k} - P_i^{m,k} \right\|,
\end{array}
\end{equation}
where $P^{n,k}$ donates the point map $P^n$ from camera $n$ transformed into camera $k$'s coordinate frame.
$\mathcal{M}$ is matching points between images \( \{I^k, I^l\} \), \( NN^{n,m}_k \) denotes the nearest neighbor matching points extracted from the point maps.
Through this, the nearest 3D point pairs can be found and used as corresponding 2D matching points. To improve matching quality, points are filtered based on the confidence map. By iterating over all pairs, all the 2D matching points $\mathcal{M}$ can be identified. 
Additionally, the inaccurate alignment in the initial point cloud also results in incorrect matches, causing failures in enforcing epipolar geometric constraints. 
Thus, we incorporate ROMA \cite{edstedt2024roma} to filter the matching points. That is to say, only matching points with high confidence values in DUSt3R results and ROMA results are preserved. The confidence thresholds of DUSt3R and ROMA are 3 and 0.05, respectively. This dual-filtering approach refines the location of 2D matching points and enables accurate bundle adjustment. 

Finally, given the 2D matching points $\mathcal{M}$, intrinsic camera matrix $\mathcal{K}$, extrinsic camera matrix $\mathcal{T}$, and the initial point cloud ${X}$, the bundle adjustment is formulate as:
\begin{equation}
\begin{array}{rl}
X^{*}, \mathcal{K}^{*}, \mathcal{T}^{*} &= \mathrm{BA} (\mathcal{M}, X, \mathcal{K}, \mathcal{T}) = \arg \min_{X, \mathcal{K}, \mathcal{T}} \mathcal{L}_{\mathrm{BA}}, \\
\mathcal{L}_{\mathrm{BA}} &= \sum_{i=1}^{N_I} \sum_{j=1}^{N_x} v_i^j \left\| \mathcal{P}_i(\mathbf{x}^j, \mathcal{K}_{i}, \mathcal{T}_{i}) - y_i^j \right\|,
\end{array}
\end{equation}
where $N_I$ is the number of input views, $N_x$ is the number of matching points, $v_i^j$ is a visibility term indicating whether point $j$ is visible in view $i$, $\mathcal{P}_i(\mathbf{x}^j,\mathcal{K}_{i},\mathcal{T}_{i})$ denotes the projection of 3D point $\mathbf{x}^j$ onto the image plane of view $i$ with camera parameter $\mathcal{K}_{i}$ and $\mathcal{T}_{i}$. $y_i^j$ is the corresponding observed 2D location of point $j$ in view $i$. 
Considering the computational complexity of global bundle adjustment, we typically limit the number of matching points selected from each view to no more than 50,000 for the calculations.

%
Upon optimizing the preserved point cloud and the camera parameters, we lower the confidence threshold for another round of bundle adjustment to optimize the position of the remaining points $X'$ with the fixed camera parameters. We can get the final point cloud $\overline{X}^*$:
\begin{equation}
\overline{X}^* = X^* \cup \mathrm{BA} (\mathcal{M}^*, X', \mathcal{K}^*,\mathcal{T}^*)
\end{equation}

\subsubsection{Semantic-based local refinement}
The point cloud after the second round of bundle adjustment remains less dense than that initially provided by DUSt3R. 
Hence we first estimate a rigid transformation $T$ to transform the initial point cloud to the optimized one:
 \begin{equation}
 T = \operatorname*{arg\,min}_T \|\overline{X}^* - TX\|.
 \end{equation}
Then, we refine the local regions with large misalignment before and after BA optimization. 
Local misaligned 3D regions are identified by measuring the point-wise distance \(\overline{X}^* - TX\) between two point clouds. These points with large errors are then clustered using DBSCAN \cite{ester1996density} and projected to 2D image views, serving as prompt points for the segmentation anything model (SAM) \cite{kirillov2023segment} in identifying local optimization areas.
For these areas, we increase the number of matching points involved in BA optimization and perform additional BA optimization to achieve a dense and accurate point cloud.


\begin{figure}[t]
    \centering
    \includegraphics[width=0.5\textwidth]{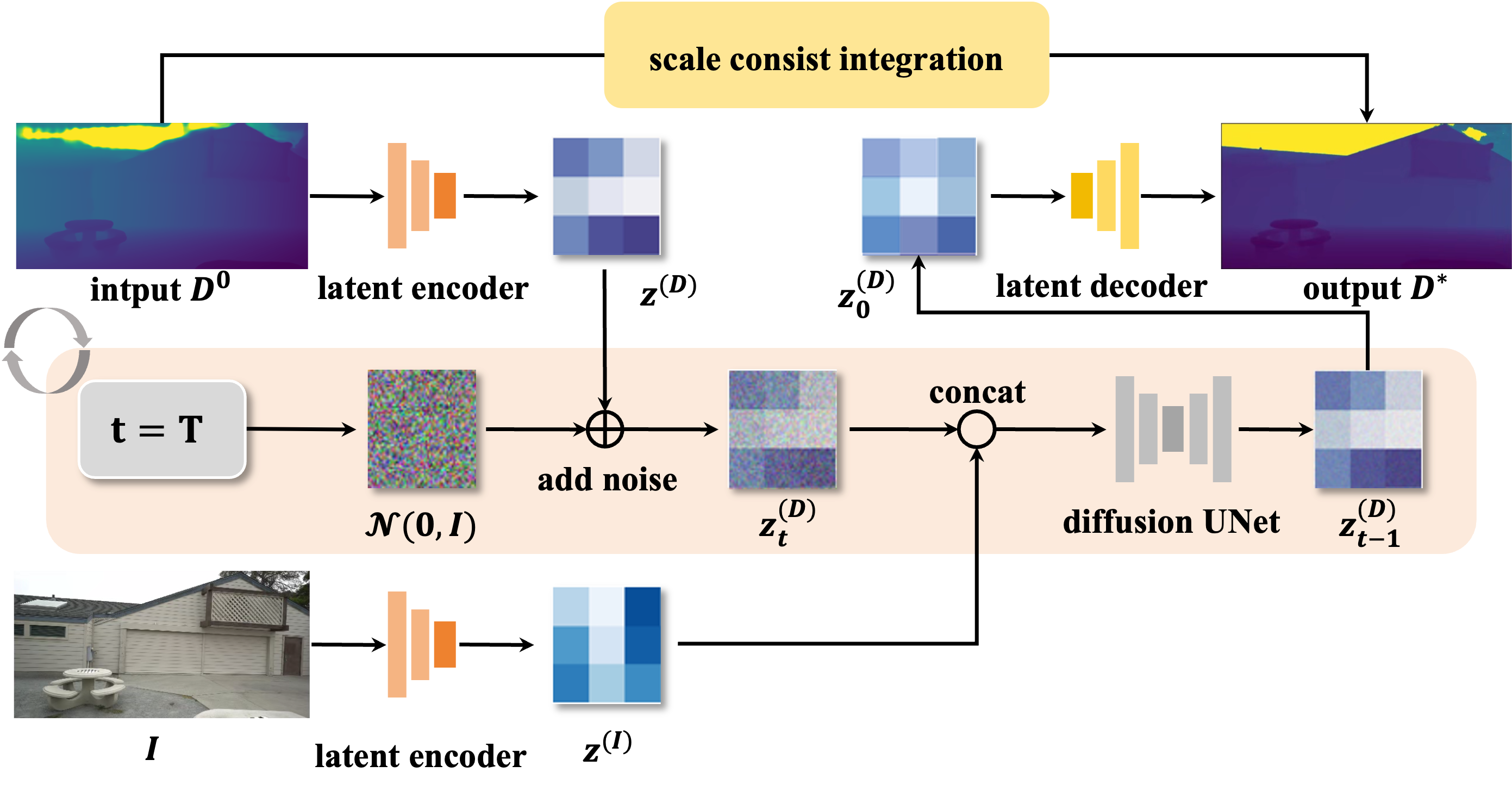} 
    \caption{Generative depth refinement module, including a diffusion model for iteratively refining the depth map and a scale-consist integration path to constrain the depth scale for each iteration. 
}
    \label{fig:generative_depth_refine}
\end{figure}

\subsection{Generative depth refinement module}
Building upon the preceding methods, we have obtained a globally accurate and dense 3D point cloud. However, the resolution limitation of the DUSt3R network (maximum \(512\times512\) pixels) restricts its ability to capture fine geometric details. Additionally, the limited overlap among sparse-view images leaves some 3D points insufficiently constrained, reducing quality in these regions.

Existing studies usually utilize the monocular depth estimation algorithms to assist sparse-view Gaussian splatting training. However, they suffer from scale ambiguity, providing only structural guidance, which is inadequate for detailed and accurate mesh reconstruction. 
Hence, we propose the generative depth refinement module, which integrates two key components: diffusion-based iterative depth refinement and scale-consistent depth integration.



\subsubsection{Diffusion-based iterative depth refinement}
Diffusion models achieve state-of-the-art results in image enhancement by learning to approximate the underlying data distribution through progressive denoising, which allows them to reconstruct images with high fidelity and fine-grained detail. Inspired by these approaches, we propose a diffusion-based algorithm to iteratively refine the depth details and accuracy.
%
As demonstrated in Fig.~\ref{fig:generative_depth_refine}, we modify Marigold’s iterative process to leverage our geometry prior. The process combines a diffusion model and scale-consistent integration, consisting of the following key steps: 
\begin{itemize}
    \item Step 1: Initial depth map projection. We project the 3D point cloud into 2D depth maps. Limited by the processing resolution of the DUSt3R network, the initial depth map $D^{0}$ usually lacks fine details.
    \item Step 2: Latent encoding. The depth map and the corresponding RGB image are input to separate latent encoders, producing latent representations \(z^{(D)}\) (depth) and \(z^{(I)}\) (RGB).
    \item Step 3: Diffusion-based refinement. 
    The latent depth representation \( z^{(D)} \) undergoes an iterative refinement process guided by a diffusion model, operating entirely in the latent code space. Starting from the initial latent depth representation \( z_0^{(D)} \), the process consists of two main stages:
    (1) In the forward process, Gaussian noise is gradually added to \( z_0^{(D)} \) over \( T \) steps to generate a sequence of noisy latent representations \( \{z_1^{(D)}, z_2^{(D)}, \dots, z_T^{(D)}\} \). At each step \( t \), this process is given by:
    \[
    z_t^{(D)} = \sqrt{\bar{\alpha}_t} z_0^{(D)} + \sqrt{1 - \bar{\alpha}_t} \epsilon,
    \]
    where \( \epsilon \sim \mathcal{N}(0, I) \), \( \bar{\alpha}_t = \prod_{s=1}^t (1 - \beta_s) \), and \( \{\beta_1, \dots, \beta_T\} \) is the noise variance schedule.
    (2) In the reverse process, the noise is progressively removed from \( z_T^{(D)} \) through \( T \) iterative steps. At each step \( t \), the diffusion UNet refines the noisy latent depth \( z_t^{(D)} \), concatenated with the RGB latent code \( z^{(I)} \), to predict and subtract the noise. After \( T \) steps, this denoising process recovers the high-quality latent depth representation \( z_0^{(D)} \).
    \item Step 4: Latent decoding. The refined depth latent is decoded into a high-resolution depth map, effectively addressing the resolution limitations of the DUSt3R network.
\end{itemize}

Our approach effectively ensures depth stability in the diffusion process while significantly improving efficiency. The original Marigold model requires 50 iterations, while our modified model only needs 10 iterations.

\subsubsection{Scale-consistent depth integration}
The diffusion process improves the depth details but may affect the depth-scale accuracy, which typically stems from two main sources: (1) The depth scale distribution gap between the training dataset of the diffusion model and our inputs. (2) Random noise introduced by the diffusion model.

To address the challenge, we add the scale-consist depth integration path to constrain the depth scale for each iteration (Fig.~\ref{fig:generative_depth_refine}). 
More specifically, we design a spatially variant depth correction algorithm using a sliding window approach: for a pixel \(i\) within a local window \( \omega_k \), the corrected depth value is $D^*_i = a_k D_i + b_k$, where \(D\) is the depth map output by diffusion process, and \(D^*\) is the corrected depth map. \(D_i\) denotes the depth value of pixel \(i\).
\( a_i \) and \( b_i \) are the coefficients obtained by minimizing the following cost function:
\begin{equation}
\begin{aligned}
E(a_i, b_i) &= \sum_{i \in \omega_k} \left( (a_i D_i + b_i - D^0_{i})^2 + \epsilon_i a_i^2 \right), \\
\epsilon_i &= 
\begin{cases} 
      \epsilon_{\text{edge}}, & \text{if } |\nabla D^0_{i}| \geq \tau_{e} \\
      \epsilon_{\text{smooth}}, & \text{if } |\nabla D^0_{i}| < \tau_{e} 
\end{cases},
\end{aligned}
\end{equation}
Where \( D^0 \) is the initial depth map projected from the DUSt3R point cloud with accurate scale, \(\nabla D^0_{i}\) is its gradient map, and \(\tau_{e} = 0.5\) is the gradient threshold. \( \epsilon_i = 10^{-7}\) is a regularization weight to prevent \( a_i \) from becoming too large, helping to control the smoothness of the depth map.
Sliding window mechanisms with windows size 25 and stride 1 and bilinear interpolation are adopted to compute the spatially variant correction parameters.
The corrected output then serves as the input for subsequent rounds of diffusion. 
To reduce variability induced by diffusion noise, we generate multiple depth maps \(D'\) through repeated inferences. We then define a set \(\mathcal{D}\) comprising depth maps that satisfy the alignment criteria:
\begin{equation}
    \mathcal{D} = \{D'\mid ||D' - D^0||_2 \leq \tau_{D} \},
\end{equation}
where \(\tau_{D} = 0.25\) is the alignment threshold.
The final depth map \(D^*\) is computed by averaging the aligned depth maps in \( \mathcal{D} \).
The resulting depth maps, which are rich in detail and accurate in scale, provide high-quality guidance for the subsequent Gaussian splatting learning and surface reconstruction. 

\subsection{multi-modal supervision}
In conventional 3D Gaussian splatting optimization, only photometric loss is considered, which usually fails for sparse-view inputs. Hence, we propose our multi-modal supervision composed of the 5 key components:
confidence-aware depth supervision, structure-aware normal supervision, synthesized pseudo-view supervision, multi-view consistency geometric supervision, and photometric loss.

\subsubsection{Confidence-aware depth supervision}
During depth optimization, some areas exhibit significant depth variation before and after optimization, resulting in lower depth confidence. To address this, we implement a confidence-aware depth loss: 
\begin{equation}
\begin{split}
\mathcal{L}_{dep} = \frac{ \sum_{x=1}^{W} \sum_{y=1}^{H} w_{\text{dep}}(x,y) \cdot |D^*(x,y) - \mathcal{D}(x,y)|}{ \sum_{x=1}^{W} \sum_{y=1}^{H} w_{\text{dep}}(x,y)},\\
 w_{dep}(x,y) = \frac{1}{1 + \beta \cdot ( |D^*(x,y) - D^{0}(x,y)|)/(|D^{0}(x,y)|)}
 \end{split}
\end{equation}
where $\beta = 0.1$ is a hyperparameter that controls the influence of depth variations and $\mathcal{D}$ is the depth map extracted from 3DGS. 
The sky region often introduces artifacts in mesh reconstruction, as the depth of the sky is almost infinite, making accurate depth estimation impractical. To solve this, we exclude sky regions from depth loss calculations. 

\subsubsection{Structure-aware normal supervision}
An accurate normal map is vital for high-quality mesh reconstruction. For each pixel in the depth map, we select its four adjacent pixels—up, down, left, and right—and project them into the 3D space, resulting in a set of 3D points $\{ \mathbf{p}_j \,|\, j = 0, \dots, 3 \}$. The normal vector is then defined as:
\begin{equation}
    \hat{\mathbf{N}}(\mathbf{p}) = \frac{(\mathbf{p}_1 - \mathbf{p}_0) \times (\mathbf{p}_3 - \mathbf{p}_2)}{|(\mathbf{p}_1 - \mathbf{p}_0) \times (\mathbf{p}_3 - \mathbf{p}_2)|}.
\label{eqn:normal_from_depth}
\end{equation}
Assume that the normal map derived from the depth map is \( \hat{\mathbf{N}} \in \mathbb{R}^{H \times W \times 3} \), and \(\hat{\mathbf{N}}(x, y)\) representing the normal vector at pixel \((x, y)\). Then its local normal consistency \( C(x, y) \) within a window is computed as:
\begin{equation}
    C(x, y) = \frac{1}{w^2} \sum_{i=-\floor{\frac{w}{2}}}^{\floor{\frac{w}{2}}} \sum_{j=-\floor{\frac{w}{2}}}^{\floor{\frac{w}{2}}} \frac{\hat{\mathbf{N}}(x + i,y + j) \cdot \hat{\mathbf{N}}_{m}}{|\hat{\mathbf{N}}(x + i,y + j)| |\hat{\mathbf{N}}_{m}|}
\end{equation}
where $w=3$ is the window size, \( \hat{\mathbf{N}}_{m} \) represents the mean normal vectors within the window. A cosine similarity $ C(x, y) $ function is then applied to convert the normal consistency from range [-1,1] to range [0,1]:
\begin{equation}
w_{nor}(x, y) = (1 + C(x, y))/2
\end{equation}
and the structure-aware normal loss is finally defined as:
\begin{equation}
\mathcal{L}_{nor} =
\frac{
    \displaystyle\sum_{x=1}^{W} \sum_{y=1}^{H} w_{nor}(x, y) \left| \mathbf{N}(x, y) - \hat{\mathbf{N}}(x, y) \right|
}{
    \displaystyle \sum_{x=1}^{W} \sum_{y=1}^{H} w_{nor}(x, y)
},
\end{equation}
where $\mathbf{N}$ denotes the normal map rendered from Gaussian splatting results.

\subsubsection{Synthesized pseudo-view supervision}
Since sparse-view inputs lack sufficient supervision from all view directions, the optimization of Gaussian primitives often converges to local minima. To address this, we generate \(n\) evenly distributed pseudo views between adjacent real views to provide additional supervision. In Gaussian primitives optimization, the weight of the virtual views is lower than that of real views.

More specifically, the pseudo-view depth maps are computed by averaging three sources:
\begin{equation}
D_t = (d_{n1} + d_{n2} + d_w)/3,
\label{eqn:depth_virtual_average}
\end{equation}
where \( d_{n1} \) and \( d_{n2} \) are the depth maps projected from the adjacent real depth maps, and \( d_w \) is the depth map projected from the point cloud.
To ensure consistency, pixels from \( d_{n1} \) or \( d_{n2} \) that differ significantly (over threshold \(\epsilon\)) from \( d_w \) are excluded from the averaging process.

The normal maps of the pseudo views can be obtained from Eq.~\ref{eqn:normal_from_depth}.  
Directly projecting the point cloud to generate pseudo-RGB views produces unsatisfactory results due to the point cloud's relative sparsity compared to image pixel density. To address this, we employ Gaussian splatting to generate pseudo-RGB views. After 3,000 training iterations, the optimized Gaussian primitives can effectively render pseudo-view RGB images.

\subsubsection{Geometric consistency loss}
Geometric consistency can be categorized into single-view and multi-view consistency. Single-view consistency ensures that the rendered normals and depth maps are geometrically consistent within a single view. Multi-view consistency enforces that the rendered depth map adheres to epipolar geometry constraints across multiple views.

For single-view consistency, we define a normal-depth consistency loss inspired by 2DGS:
\begin{equation}
\mathcal{L}_{ndc} = \frac{1}{WH} \sum |\nabla D| \| \hat{\textbf{N}} - \textbf{N} \|_1,
\label{eq:loss_svggeo}
\end{equation}
where \( \nabla D \) is the depth gradient normalized to the range [0, 1], \( \hat{\textbf{N}} \) is given by Eq.~\ref{eqn:normal_from_depth}, and \(\textbf{N}\) denotes the normal map rendered from Gaussian
splatting results. The depth gradient \( \nabla D \) is used as a weighting factor in the loss function, emphasizing consistency in edge geometry.

For evaluating multi-view consistency, we propose a cycle-projection loss:  
pixels \( \{x, y\} \) from an original view are first projected into 3D space using the rendered depth map, then reprojected into the 2D space of a neighboring view. A backward projection is then performed using the neighboring rendered depth map to project the pixels back to the original view, obtaining \( \{x', y'\} \). By comparing the distances, we define the multi-view consistency loss:
\begin{equation}
\mathcal{L}_{mv} = \frac{1}{WH} \sum_{x}^{W} \sum_{y}^{H} \left\| x - x' \right\|^2 + \left\| y - y' \right\|^2
\end{equation}
The projection and back-projection matrices can be derived from the camera's intrinsic and extrinsic parameters of the two views.

For pseudo views, the neighboring view is defined as the closest real view, as the geometry of real views is typically more accurate. For real views, two options are considered: one involves randomly generating a nearby view to perform the forward and backward projections and calculate consistency, while the other involves using the nearest real view to compute geometric consistency. These two options are selected randomly.

\subsubsection{Photometric loss}
The photometric loss function incorporates both pixel-wise intensity differences and structural similarity:
\begin{equation}
    \mathcal{L}_{pho} = (1 - \lambda) \mathcal{L}_1 + \lambda \mathcal{L}_{\text{SSIM}},
\end{equation}
\( \lambda=0.2 \) is a hyperparameter that adjusts the relative importance of the two terms. \(\mathcal{L}_1\) and \(\mathcal{L}_{\text{SSIM}}\) are the \(L_1\) loss and SSIM loss, respectively.

Finally, all losses are integrated as follows:
\begin{equation}
\mathcal{L} = \lambda_{1}\mathcal{L}_{nor} +  \lambda_{2}\mathcal{L}_{dep} + \lambda_{3}\mathcal{L}_{ndc} + \lambda_{4}\mathcal{L}_{mv} + \mathcal{L}_{pho}.
\end{equation}
We set $\lambda_{1}=0.005$,$\lambda_{2}=0.005$,$\lambda_{3}=0.1$,and $\lambda_{4}=0.1$. 
Upon completing the training of the Gaussian primitives, we utilize the TSDF fusion method \cite{curless1996volumetric} to merge the depth maps from all views, resulting in the final mesh.

%% file: exp.tex
\begin{figure*}[t]
    \centering
    \includegraphics[width=0.95\textwidth]{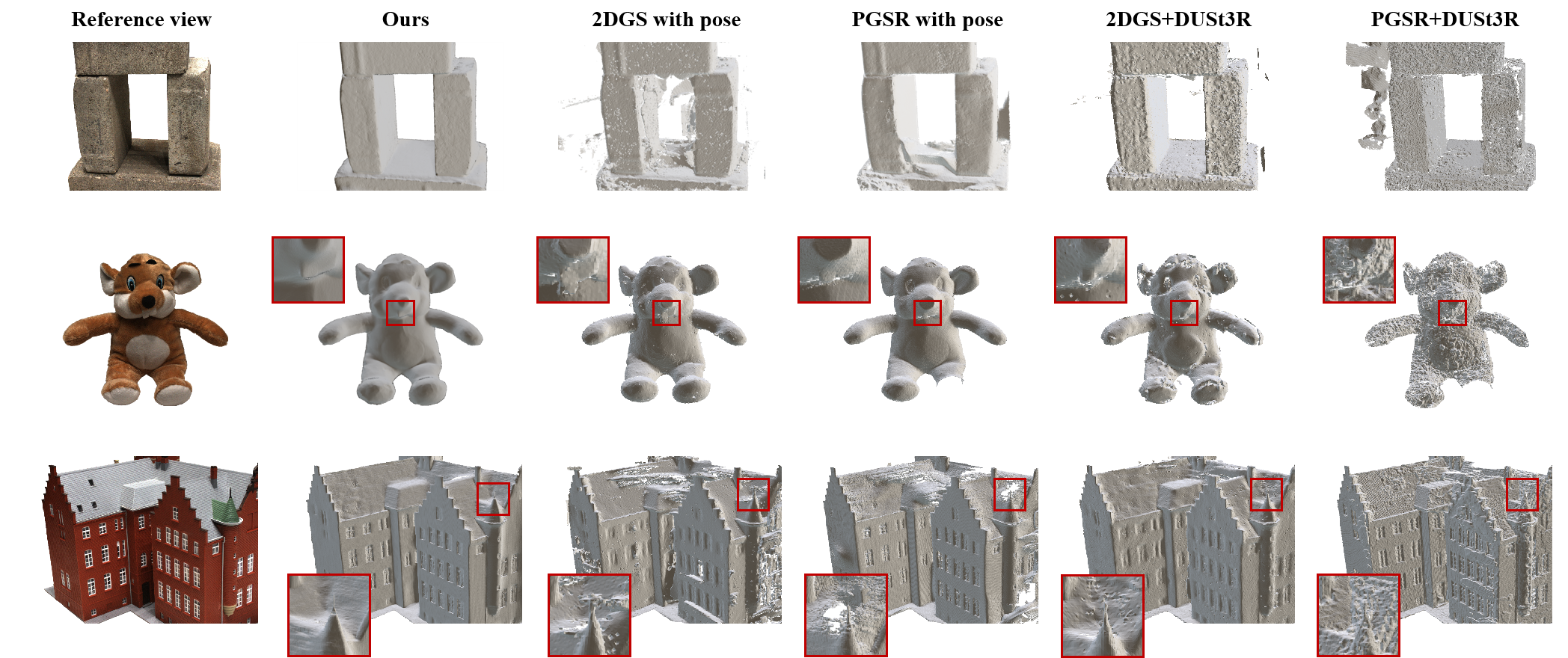} 
    \caption{Qualitative results using 4 input views on DTU dataset. Compared to DUSt3R initialization, our method achieves superior geometric accuracy. Compared to baseline methods using accurate poses computed by COLMAP, our approach generates more complete surface meshes.}
    \label{fig:exp_dtu}
\end{figure*}
\section{Experiments}
\subsection{Dataset and sparse view selection}

\begin{figure*}[t]
    \centering
    \includegraphics[width=0.95\textwidth]{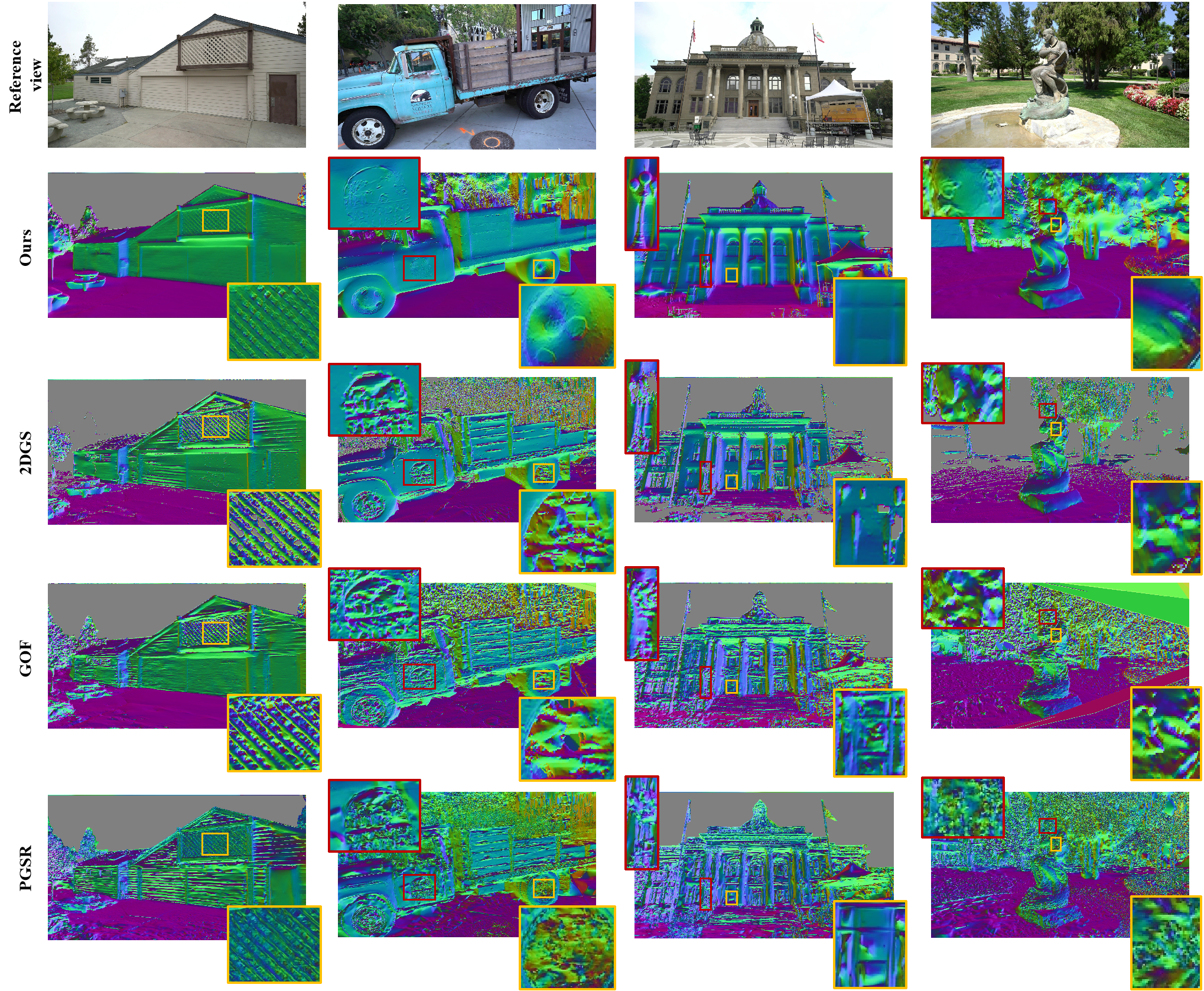} 
    \caption{Qualitative results using 6 input views on Tanks \& Temples dataset. The figure shows surface reconstruction quality using normal maps, with sky regions masked out. Our method generates smooth, detail-rich geometry, outperforming others that struggle with noise and fail to capture fine details. Additionally, our approach reconstructs both foreground and background geometry, unlike other methods restricted to foreground objects.}
    \label{fig:exp_tnt}
\end{figure*}

\begin{figure*}[t]
    \centering
    \includegraphics[width=1.0\textwidth]{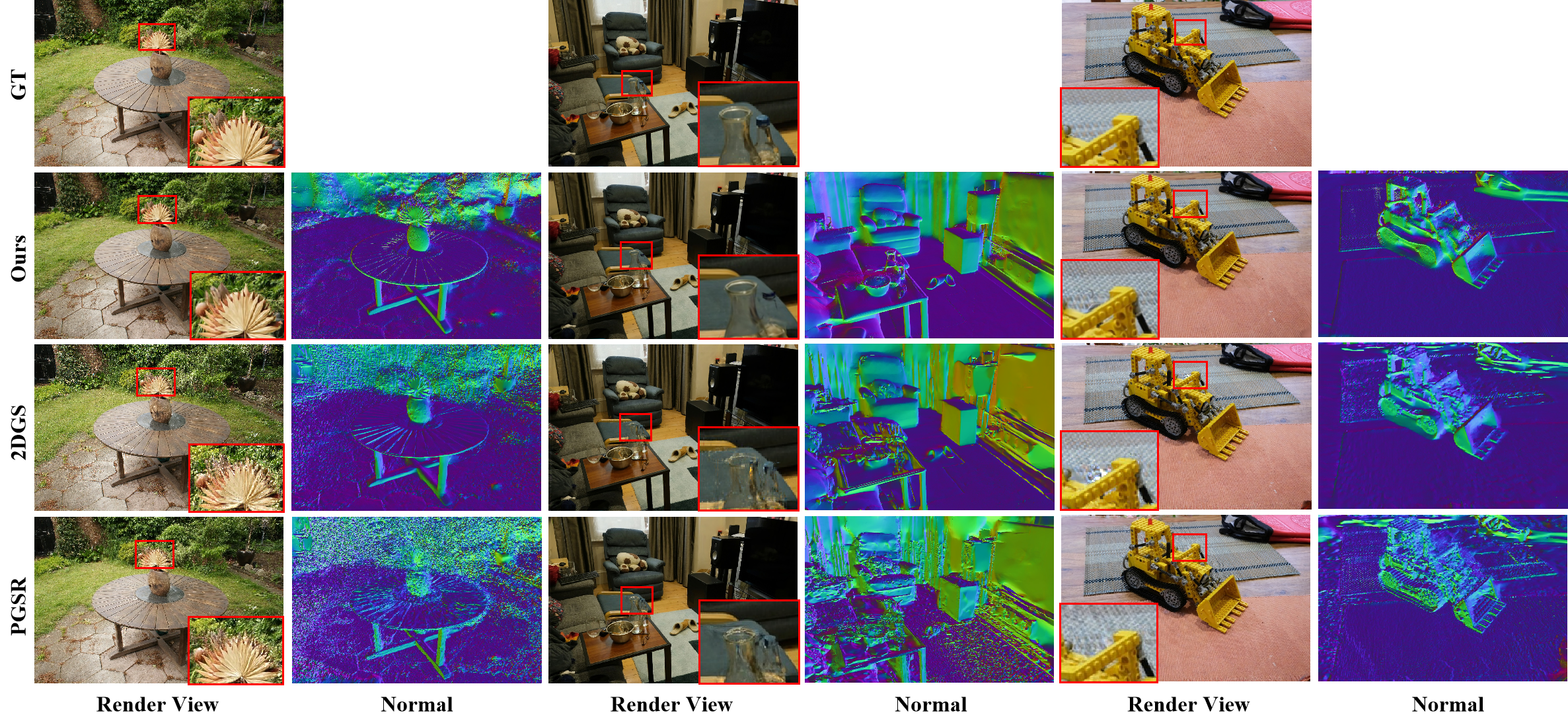} 
    \caption{
    Qualitative NVS results using 6 input views on the Mip-NeRF 360 dataset. The figure demonstrates the synthesized RGB images and normal maps of different methods. Our approach renders novel views with sharper details and fewer artifacts.}
    \label{fig:exp_mip360}
\end{figure*}

Similar to previous algorithms, we selected the DTU \cite{aanaes2016large}, Tanks \& Temples \cite{knapitsch2017tanks}, and Mip-NeRF 360 \cite{barron2022mip}  datasets for testing our method. The DTU and TNT datasets are primarily used to evaluate geometric reconstruction performance, while the Mip-NeRF 360 dataset is employed to assess the algorithm's ability to generate novel views and to qualitatively showcase the mesh reconstruction results. 
For the DTU dataset, we selected 4 views for training and testing. This sparse-view configuration presents a significant challenge. For the larger scene datasets (Tanks \& Temples and Mip-NeRF 360), we used 6 views as training data and an additional 6 neighboring views for testing.
For the DTU and Tanks \& Temples datasets, we sampled the ground truth 3D geometry and only preserved those points visible from two or more views. 
For the Tanks \& Temples dataset testing, irrelevant point clouds, such as sky regions and background vegetation, were excluded to improve the quality of the results.

In addition, to further validate the effectiveness of our algorithm, we conducted experiments on the large-scale GigaNVS dataset \cite{Wang_2024_CVPR}, utilizing only six images per scene. The qualitative results on GigaNVS are shown in teaser and supplementary material.

\subsection{Evaluation criteria}
For the novel view synthesis task, we employed three widely used image evaluation metrics: Peak Signal-to-Noise Ratio (PSNR), Structural Similarity Index Measure (SSIM), and Learned Perceptual Image Patch Similarity (LPIPS).
For surface quality evaluation, we assessed performance using the \(F_1\) score (Tanks \& Temples) and Chamfer Distance (DTU). Additionally, we report the Absolute Trajectory Error (ATE) to evaluate the effectiveness of the camera parameters optimization. The real camera poses provided by the dataset are used as the ground truth.

\subsection{Implementation details}
In the context of BA, optimizing the camera's intrinsic and extrinsic parameters, as well as a subset of the point cloud, requires the selection of high-confidence point clouds for the calculations. Therefore, when the number of input views exceeds three, we set the confidence threshold to 3, meaning that a 3D point must be observed by at least three views. In other scenarios, the confidence threshold is set to 2. For point cloud filtering, we set the confidence threshold to 3.

For the training of Gaussian primitives, the basic training process and hyperparameters are similar to those of the original 3D Gaussian Splatting. However, a notable difference is that, due to the use of sufficiently dense point clouds for initialization, the number of training iterations can be significantly reduced. Specifically, we first train for 3,000 iterations, during which satisfactory results are often achieved. Subsequently, we render the RGB images for the pseudo views and perform depth optimization for these views before training for another 3,000 iterations. Comparatively, the conventional 3D Gaussian splatting typically requires around 300,000 iterations.

Generally, we adopt the same densification strategy as AbsGS \cite{ye2024absgs}. However, due to the density of the initial point cloud, we do not apply the strategy during the first 1,500 iterations. The frequency of applying densification is also adjusted in subsequent iterations, occurring once every 500 iterations. All experiments were conducted using an NVIDIA RTX 4090 GPU.

\subsection{Baselines}
We combine DUSt3R \cite{wang2024dust3r} with other Gaussian splatting-based surface reconstruction algorithms as the primary baseline. For the optimization of DUSt3R, we employ the method of InstantSplat \cite{fan2024instantsplat} to average the intrinsic parameters, which can improve the accuracy of the generated 3D point cloud.
%
%
Additionally, we also conducted experiments using COLMAP combined with the Gaussian splatting-based surface reconstruction algorithms. However, this approach did not yield satisfactory results on the Tanks \& Temples dataset and is not included in our evaluations.
For surface reconstruction algorithms, we select 2DGS \cite{huang20242d}, GOF \cite{Yu2024GOF}, and PGSR \cite{chen2024pgsr}. We excluded SuGaR \cite{guedon2024sugar} from our evaluation due to its relatively high failure rate.

\begin{table*}[ht]
\centering
\renewcommand{\arraystretch}{1.3}
\setlength{\tabcolsep}{4pt}
\begin{tabular}{l|ccccccccccccccc|c} 
\toprule
\textbf{Method} & \textbf{24} & \textbf{37} & \textbf{40} & \textbf{55} & \textbf{63} & \textbf{65} & \textbf{69} & \textbf{83} & \textbf{97} & \textbf{105} & \textbf{106} & \textbf{110} & \textbf{114} & \textbf{118}& \textbf{122}& \textbf{Mean}  \\
\midrule
2DGS+DUSt3R &0.833 & 1.475 & 1.759 & 0.995 & 1.779 & 1.598 & 1.460 & 1.649 & 1.805 & 1.986 & 1.546 & 1.943 & 1.296 & 1.419 & 1.289 & 1.522\\
PGSR+DUSt3R  &0.950 & 1.685 & 1.322 & 1.028 & 1.545 & 1.823 & 1.366 & 1.921 & 1.793 & 1.386 & 1.379 & 2.387 & 1.154 & 1.186 & 1.140 & 1.471 \\
GOF+DUSt3R   & 1.098 & 1.825 & 2.130 & 1.196 & 1.766 & 2.077 & 1.608 & 1.880 & 1.997 & 1.935 & 1.862 & 1.892 & 1.347 & 2.072 & 2.721 & 1.827 \\
2DGS+COLMAP & 1.327 & 0.877 & 1.239 & 0.565 & 1.008 & 0.953 & 1.365 & 1.020 & 1.453 & 1.053 & 1.268 & - & 0.676 & 1.406 & 1.670 & 1.134 \\
PGSR+COLMAP  &1.511 & 5.176 & 1.530 & \cellcolor{red!20}0.364 & \cellcolor{red!20}0.805 & \cellcolor{red!20}0.702 & \cellcolor{red!20}0.967 & \cellcolor{red!20}0.641 & 1.349 & \cellcolor{red!20}0.887 & \cellcolor{red!20}0.669 & - & 0.584 & \cellcolor{red!20}0.628 & \cellcolor{red!20}1.181 & 1.214 \\
GOF+COLMAP   &\cellcolor{red!20} 0.604 &\cellcolor{red!20} 0.779 & \cellcolor{red!20}1.110 & 0.415 & 1.138 & 0.781 & 1.055 & 0.863 & \cellcolor{red!20}1.003 & 0.986 & 0.942 & - & \cellcolor{red!20}0.493 & 0.768 & 3.251 & \cellcolor{red!20}1.013
 \\
Ours & \cellcolor{red!50}\textbf{0.413} & \cellcolor{red!50}\textbf{0.642} & \cellcolor{red!50}\textbf{0.555} & \cellcolor{red!50}\textbf{0.281} & \cellcolor{red!50}\textbf{0.505} & \cellcolor{red!50}\textbf{0.553} & \cellcolor{red!50}\textbf{0.525} & \cellcolor{red!50}\textbf{0.611} & \cellcolor{red!50}\textbf{0.667} & \cellcolor{red!50}\textbf{0.587} & \cellcolor{red!50}\textbf{0.406} & \cellcolor{red!50}\textbf{0.664} & \cellcolor{red!50}\textbf{0.393} & \cellcolor{red!50}\textbf{0.535} & \cellcolor{red!50}\textbf{0.561} & \cellcolor{red!50}\textbf{0.526} \\

\bottomrule
\end{tabular}
\caption{Quantitative comparison of methods on DTU data (Chamfer distance mm $\downarrow$). Red cells indicate the best performance, and light red cells indicate the second-best performance for each metric. `-` indicates experiment failure.}
\label{tab:DTU_comparison}
\end{table*}

\begin{table}[ht]
\centering
\setlength{\tabcolsep}{4pt} 
\renewcommand{\arraystretch}{1.3} 
\begin{tabular}{lcccc}
\toprule
 & 2DGS & GOF & PGSR & Ours \\
\midrule
Barn & \cellcolor{red!20}0.1855 & 0.1517 &0.1731 & \cellcolor{red!50}\textbf{0.2965} \\
Caterpillar &\cellcolor{red!20}0.0808&0.0323&0.0755& \cellcolor{red!50}\textbf{0.1998}  \\
Courthouse &  0.0446 & 0.0385 &\cellcolor{red!20}0.1112& \cellcolor{red!50}\textbf{0.1452} \\
Ignatius &0.0636 &0.0685&\cellcolor{red!20}0.1315 & \cellcolor{red!50}\textbf{0.3126} \\
Meeting room & 0.0314 & - &\cellcolor{red!20}0.0399& \cellcolor{red!50}\textbf{0.1227} \\
Truck & \cellcolor{red!20}0.1503 & 0.1183 &0.1027& \cellcolor{red!50}\textbf{0.2562} \\
\midrule
Mean &0.0927&0.0682& \cellcolor{red!20}0.1057 & \cellcolor{red!50}\textbf{0.2217} \\

\bottomrule
\end{tabular}
\caption{Quantitative result of $F_1$ score$\uparrow$ on Tanks and Temples dataset.}
\label{tab:exp_tnt}
\end{table}

\begin{table}[ht]
\centering
\setlength{\tabcolsep}{10pt} 
\renewcommand{\arraystretch}{1.3} 
\begin{tabular}{lcccc}
\toprule
\textbf{Method} & \textbf{PSNR$\uparrow$} & \textbf{SSIM$\uparrow$} & \textbf{LPIPS$\downarrow$} & \textbf{ATE$\downarrow$} \\
\midrule
2DGS & 19.831 & 0.482 & 0.358 &0.222\\ 
GOF & \cellcolor{red!20}20.967 & \cellcolor{red!20}0.530 & \cellcolor{red!20}0.310 &0.222\\ 
PGSR & 19.613 & 0.491 & 0.347 &0.222\\ 
InstantSplat &18.755 & 0.510 &0.420 &\cellcolor{red!20}0.126 \\
CF-3DGS&15.516&0.2972&0.5508&0.275\\
Ours & \cellcolor{red!50}\textbf{23.363} &  \cellcolor{red!50}\textbf{0.658} &  \cellcolor{red!50}\textbf{0.241} &\cellcolor{red!50}\textbf{0.027}\\

\bottomrule
\end{tabular}
\caption{Quantitative comparison of NVS task on Mip-NeRF 360 dataset. PSNR and SSIM are higher the better, while LPIPS is lower the better.}
\label{tab:exp_mip360}
\end{table}

\begin{table}[ht]
\centering
\setlength{\tabcolsep}{10pt} 
\renewcommand{\arraystretch}{1.3} 
\begin{tabular}{lcccc}
\toprule
\textbf{Method} & \textbf{3 views} & \textbf{4 views} & \textbf{5 views} & \textbf{6 views} \\
\midrule
2DGS+DUSt3R & 1.589 & 1.522 &  1.454 &1.438\\ 
PGSR+DUSt3R & 1.478 & 1.471 & 1.464 &1.358\\ 
Ours & \cellcolor{red!50}\textbf{0.543} &  \cellcolor{red!50}\textbf{0.526} &  \cellcolor{red!50}\textbf{0.520} &\cellcolor{red!50}\textbf{0.521}\\

\bottomrule
\end{tabular}
\caption{Quantitative comparison of different methods on the DTU dataset (Chamfer distance mm $\downarrow$) under varying numbers of input views.}
\label{tab:exp_num_view}
\end{table}

\begin{table}[ht]
\centering
\setlength{\tabcolsep}{4pt} 
\renewcommand{\arraystretch}{1.3} 
\begin{tabular}{lcccc}
\toprule
 & precision & recall & f-score \\
\midrule
w/o Neural-BA  & 0.2287 & 0.2216 &0.2251 \\
w/o Depth Refinement &0.2836&0.2874& 0.2855  \\
w/o Depth Loss &0.2461&0.2520&0.2490  \\
w/o Geometric consistency Loss &0.2786&0.2622&0.2701  \\
w/o Pseudo views &0.2742&0.2791&0.2766  \\
Full&\textbf{0.2975}&\textbf{0.2954}&\textbf{0.2965}\\

\bottomrule
\end{tabular}
\caption{Quantitative result of $F_1$ score$\uparrow$ on Barn scene of Tanks \& Temples dataset.}
\label{tab:exp_albation}
\end{table}

\subsection{Geometric Evaluation on the Synthetic Dataset}
Table \ref{tab:DTU_comparison} presents the quantitative results of the geometric evaluations for the synthetic DTU dataset. 
As discussed earlier, network-based geometry estimation methods, such as DUSt3R, suffer from low accuracy. While the conventional COLMAP method can produce highly accurate results, it is less robust for sparse-view inputs. For instance, COLMAP fails to reconstruct the 100 data with only 4-view inputs.
While our method combines the advantages of these two methods, significantly outperforming the baseline methods in most cases. Only in a few specific scenarios, such as scenes 55, 93, and 114, where COLMAP can provide accurate extrinsic parameters, the baseline methods show marginal competitiveness with our method.

As qualitatively shown in Fig.~\ref{fig:exp_dtu}, COLMAP struggles to produce a complete mesh, resulting in significant holes. DUSt3R, on the other hand, generates an inaccurate initial point cloud and camera parameters, which leads to a noisy and non-smooth mesh. In contrast, our method ensures both high accuracy and a dense initial point cloud. It remains robust even with extremely sparse viewpoints, yielding accurate, high-fidelity mesh reconstruction.

\subsection{Geometric Evaluation on Real-world Large-scale Scenes}
Compared to the DTU dataset, the Tanks \& Temples dataset collects outdoor real-world data, which presents more significant challenges.
The numerical evaluation results in Table \ref{tab:exp_tnt} demonstrate that our method holds a clear advantage across all scenes.
Fig.~\ref{fig:exp_tnt} shows the rendered normal maps. Our normal map is more complete and cleaner than the baseline methods, with better object details. For example, in our results, the eyes, nose, and fingers of the sculpture are successfully reconstructed, while the other methods fail to generate reasonable meshes and normal maps.


\subsection{Evaluation of Novel View Synthesis and Camera Pose Estimation}
We use the Mip-NeRF 360 dataset to test the performance of novel view synthesis and pose estimation. In addition to the baseline methods, we also evaluated two pose-free methods: InstantSplat and COLMAP-free 3D Gaussian splatting (CF-3DGS) \cite{Fu2023COLMAPFree3G}. 
InstantSplat uses DUSt3R for initialization but without extensive optimization, simply setting the focal length to an average value.
CF-3DGS processes the input frames in a sequential manner and progressively grows the 3D Gaussians set by taking one input frame at a time. 

As shown in Table \ref{tab:exp_mip360}, our method outperforms the others in image rendering quality, and also estimates more accurate camera poses (ATE). Notably, PSNR improves by more than 2 dBs.
The qualitative results, presented in Fig.~\ref{fig:exp_mip360}, demonstrate that our method generates cleaner and smoother normal maps, leading to artifact-free, high-quality novel view synthesis results.
Compared to the original DUSt3R (InstantSplat), the ATE error is reduced by 20\%, significantly improving pose accuracy and enhancing the quality of surface reconstruction.


\begin{figure}[t]
    \centering
    \includegraphics[width=0.5\textwidth]{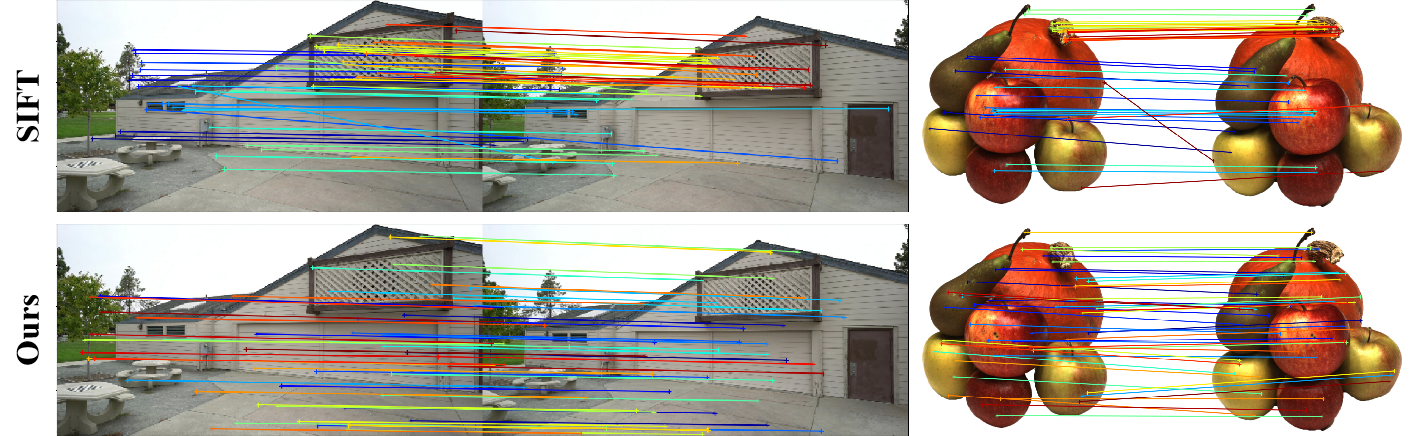} 
    \caption{
    Qualitative comparison of feature extraction and matching results. Compared to traditional feature matching methods, our approach effectively handles texture-less regions, achieving a more uniform distribution of keypoints across the scene and improving correspondence reliability.}
    \label{fig:exp_feature}
\end{figure}

\subsection{Robustness Testing}
\subsubsection{Robustness of Feature Point Matching}
Unlike traditional methods that rely on local image content features, which often struggle in texture-less regions, our algorithm utilizes the powerful feature extraction capabilities of transformers. As shown in Fig.~\ref{fig:exp_feature}, by emphasizing global image-level representations over localized details, our method effectively handles texture-less regions, such as floors, leaves, fruit skins, as well as other large and uniform surfaces. It ensures a more uniform distribution of keypoint correspondences across the scene, enabling more reliable and accurate matching in these challenging areas.

\subsubsection{Robustness Across Varying Numbers of Views}
To further demonstrate the impact of view count on algorithm performance, we supplemented our analysis on the DTU dataset with an additional set of experiments. Specifically, we evaluated the performance under three-view settings for increased difficulty, the existing four-view setting, five-view setting and a six-view setting comparable to the configuration used for Tanks \& Temples. The experimental results in Table \ref{tab:exp_num_view} show that our algorithm consistently outperforms others across varying numbers of views, demonstrating its robustness to sparse-view conditions.

\subsection{Ablations}
Table \ref{tab:exp_albation} presents the results of our ablation study, which evaluates the contribution of the neural-BA module, depth refinement module, depth loss, geometric consistency loss, and pseudo view loss. The Barn scene from the Tanks \& Temples dataset is used for testing.
Each row corresponds to a different configuration of the system, with one key component excluded. Precision, recall, and $F_1$ score are used to show the impact of all the components. The results effectively highlight the critical role of all the components in the final performance.

\section{Limitations}
Our method encounters challenges when dealing with scenes involving specular reflections and transparent objects. Addressing this limitation may require incorporating polarized imaging data or training advanced foundation models to better recognize such objects. Additionally, extending the method to support 4D reconstruction could be achieved in the future by integrating a motion-aware module.

\section{Conclusion}
In this paper, we presented GBR: Generative Bundle Refinement, a framework designed for high-fidelity Gaussian splatting and meshing. By integrating neural bundle adjustment, generative depth refinement, and a multimodal loss function, GBR significantly improves geometric accuracy, detail preservation, and robustness in Gaussian splatting optimization.
Our method effectively recovers camera parameters, generates dense point maps, and supports novel view synthesis and mesh reconstruction from sparse input images. We validated the performance of GBR on the DTU, Tanks \& Temples, and Mip-NeRF 360 datasets, demonstrating its superiority over state-of-the-art methods in both geometric accuracy and novel view rendering quality under sparse-view conditions. Finally, we demonstrated GBR's applicability to large-scale real-world scenes, such as the Pavilion of Prince Teng and the Great Wall.